\documentclass{article} %
\usepackage{iclr2027_conference,times}

\usepackage{amsmath,amsfonts,bm}

\def\eqref#1{equation~\ref{#1}}

\def\1{\bm{1}}

\DeclareMathAlphabet{\mathsfit}{\encodingdefault}{\sfdefault}{m}{sl}
\SetMathAlphabet{\mathsfit}{bold}{\encodingdefault}{\sfdefault}{bx}{n}

\usepackage{hyperref}
\usepackage{url}
\usepackage{todonotes}
\usepackage{pgfplots}
\usepackage{pgfplotstable}
\usepackage{enumitem}
\usepackage{tabularx}
\usepackage{booktabs}
\usepackage{amsthm}
\usepackage{caption}
\usepackage{makecell}
\usepackage{wrapfig}
\usepackage{subcaption}
\newtheorem{proposition}{Proposition}

\pgfplotsset{compat=1.18}
\definecolor{ponderblue}{HTML}{0072B2}
\definecolor{baselineorange}{HTML}{D55E00}
\pgfplotsset{
    scaling axis/.style={
        width=\linewidth,
        height=3.55cm,
        ymajorgrids=true,
        grid style={draw=gray!18},
        axis line style={draw=gray!55},
        tick style={draw=gray!55},
        tick label style={font=\scriptsize},
        label style={font=\scriptsize},
        xlabel={Rethinking budget $R$},
        unbounded coords=discard,
    },
    ponder scaling/.style={
        ponderblue, thick, mark=*, mark size=1.5pt,
    },
    random scaling/.style={
        baselineorange, densely dashed, thick,
        mark=square*, mark size=1.5pt,
    },
    runtime axis/.style={
        scaling axis,
        height=4.3cm,
        xlabel={Time per sample (ms)},
        tick pos=left,
        legend style={font=\tiny, draw=none, fill=none, row sep=-2pt},
        legend cell align=left,
        nodes near coords style={font=\tiny, text=black},
    },
}

\newcommand{\method}{\textsc{Ponder}}

\title{Learning to Generate in the Right Order}

\author{%
\begin{minipage}{0.97\textwidth}
\centering
\normalfont
\vspace{9pt}
\textbf{%
Julius Erbach\textsuperscript{$\dagger$,$\ddagger$}\quad
Dominik Narnhofer\textsuperscript{$\dagger$}\quad
Bernt Schiele\textsuperscript{$\ddagger$}}\\[4pt]
\textbf{%
Jan Eric Lenssen\textsuperscript{$\ddagger$,\P}\quad
Konrad Schindler\textsuperscript{$\dagger$}}\\[6pt]
{\small
\textsuperscript{$\dagger$}ETH Zurich\quad
\textsuperscript{\P}Saarland University\\[2pt]
\textsuperscript{$\ddagger$}Max Planck Institute for Informatics, Saarland Informatics Campus
}
\end{minipage}%
}

\iclrfinalcopy %
\begin{document}

\maketitle
\lhead{Preprint}
\vspace{-1.em}

\begin{abstract}
\vspace{-.5em}

We introduce \method{}, a method that learns in which order an image generator should fill in content and which of its earlier predictions to revise.
Masked generative models construct a sample by repeatedly filling in subsets of missing variables. The order of these partial predictions determines the context available in subsequent steps and therefore affects the quality of the final sample. Common strategies are to choose the order randomly or based on local confidence estimates. 
Conditioned on the partially generated sample and the generator’s features, our policy chooses which positions to fill next and/or which previously filled positions to mask for later regeneration, a mechanism we call rethinking. In this way, the generation process is able to adapt to the evolving context and also to revise earlier predictions. We optimize the policy using group-relative policy gradients, with rewards that measure constraint satisfaction, human preference, or distribution quality. For the latter, we introduce FIDgrad, an image-level reward derived from the (first-order) change in Fréchet Inception Distance caused by adding a generated sample to the current feature distribution. With the generators frozen, learned ordering reduces MAR-B ImageNet FID from 4.09 to 2.78, improves MaskGen-KL-XL HPSv2 from 25.05 to 26.50, and raises the accuracy on hard visual Sudokus from 4.5\% to 56.6\%, surpassing uncertainty-based greedy ordering (41.0\%). Rethinking further improves these results to 2.70 FID, 27.80 HPSv2 and 96.7\% accuracy. Our code is available \href{https://juliuserbach.github.io/Ponder/}{here}.
\end{abstract}

\begin{figure}[!htbp]
    \centering
    \input{figures/teaser.tex}
    \vspace{-1em}
    \caption{\textbf{Learning to generate and rethink.} Left: a learned policy selects positions of patches to generate or mask for later regeneration, enabling revision of earlier predictions. Right: increasing the inference-time rethinking budget improves Sudoku-hard accuracy and MAR-B image quality, with each policy and generator held fixed.}\label{fig:teaser}
\end{figure}

\section{Introduction}
Standard diffusion models generate images by iteratively denoising all spatial locations jointly~\citep{hoDenoisingDiffusionProbabilistic2020, lipmanFlowMatchingGenerative2023a, karrasElucidatingDesignSpace2022a, rombach_high-resolution_2022}. %
An alternative is to factorize generation over spatial locations, using a diffusion model to predict individual patches or subsets conditioned on content that has already been generated~\citep{li_autoregressive_2024}. 
In this setting, the generation order becomes consequential because each prediction determines the context available for subsequent ones.
It has been shown that such conditional generation is capable of high-quality image synthesis and spatial reasoning, but also that its performance depends substantially on the chosen patch ordering~\citep{fan_fluid_2024, wewer25srm}.

If all conditionals were known exactly, different orderings would yield equivalent factorizations. However, the conditionals that are actually learned do not preserve the equivalence. In practice, generation order substantially affects sample quality~\citep{wewer25srm, li_autoregressive_2024, fan_fluid_2024}. This observation leads to a natural question: \emph{Can we learn to adapt the ordering in a sample-dependent way, such that the complete, final sample has the best possible quality?}

There are three common strategies to define the order: \emph{Fixed} schedules, most often in raster-scan order, are independent of the generated content~\citep{esser2021taming, wang_emu3_2024, zai2026glmimage}. \emph{Random} schedules sample a permutation per roll-out, as in masked autoregressive models~\citep{li_autoregressive_2024}. These decorrelate training from any particular order but provide no mechanism for exploiting context during inference. \emph{Greedy} schedules prioritize positions according to confidence or (predicted) difficulty: MaskGIT~\citep{changMaskGITMaskedGenerative2022, li_nemotron-labs-diffusion-image_2026} unmasks the tokens the generator is most confident about, SRM~\citep{wewer25srm} selects the ones with the lowest predicted uncertainty, and Patch Forcing~\citep{schusterbauer2026patchforcing} estimates a per-patch difficulty score and advances easy regions first, so as to obtain more context for harder ones.
All three families share the same limitation: either the order of generation is completely independent of the sample being generated, or it is determined by a greedy one-step proxy.%

Among those strategies, only greedy confidence-based schedules adapt the ordering to the intermediate, partial state at all. However, the decision is based on an extremely local proxy: it prioritizes positions that are easy to predict given the current context. Such a strategy does not consider how complex future steps will become based on that decision. Thus, a local prediction that is made with high confidence does not, by itself, mean that it will greatly improve the global sample quality. A highly predictable position may provide little useful context, whereas a less certain one may establish useful structure that constrains further patches.

The delayed effect of intermediate decisions turns generative modeling into a sequential decision-making problem, driven by a policy that selects positions conditioned on the current partial sample. Taking that view further, earlier predictions made with limited context may become inconsistent at later stages of the generation process, calling for a mechanism to also revisit earlier choices. 

We introduce \textbf{\method{}} (Policy for Ordering, Next-element DEcisions and Rethinking), a scheme for learning a policy that controls the sequence order during generation, including removal of earlier predictions for subsequent regeneration. To that end, we make the following contributions:
\begin{itemize}[itemsep=4pt, topsep=2pt, parsep=2pt]
    \item \emph{Learned ordering for continuous generators.} A policy trained with final-sample rewards that decides where a frozen generator fills content next. It requires only the policy's own action probabilities, not the generator's likelihoods, which makes it the first such approach applicable to continuous-token generators with diffusion heads. 

    \item A \emph{rethinking} strategy, where the policy
    learns which earlier predictions to discard and regenerate as
    new context becomes available. This enables adaptive revision
    during sequential generation with a controllable rethinking budget that can be varied at inference time.

    \item \emph{FIDgrad}, a reward function that turns a distribution-level objective into an image-level reinforcement learning signal. By estimating each sample's first-order effect on Fréchet Inception Distance, FIDgrad enables a generation policy that directly targets quality at the distribution level.
\end{itemize}
A suite of experiments demonstrates the generality of \method{} across class-conditional generation, text-to-image synthesis, and visual constraint solving. With learned ordering alone, it reduces MAR-B ImageNet FID from 4.09 to 2.78, improves MaskGen-KL-XL HPSv2 from 25.05 to 26.50, and raises hard-Sudoku accuracy from 4.5\% to 56.6\%. Rethinking further improves these results, to 2.70 FID, 27.80 HPSv2 at $R=32$, and 96.7\% accuracy. Moreover, performance continues to improve as the inference-time budget increases while the policy and generator remain fixed (Figure~\ref{fig:teaser}).

\section{Related Work}

\paragraph{Masked generation with continuous representations.}
Masked generative models predict missing content from a partially observed sample, allowing generation to proceed in different orders autoregressively. MaskGIT~\citep{changMaskGITMaskedGenerative2022} uses discrete image tokens and iterative confidence-based decoding. MAR~\citep{li_autoregressive_2024} instead models conditional distributions over continuous patches using a patch-wise diffusion head, removing the need for vector quantization. FLUID~\citep{fan_fluid_2024} studies the effects of token representation and generation order in text-to-image synthesis, finding advantages for continuous tokens and random-order generation. MaskGen~\citep{maskgen} supports both discrete and continuous representations through compact one-dimensional image tokens. We build on continuous masked generators and learn how to use their conditional predictions while keeping their weights frozen.

\paragraph{Adaptive and learned generation orders.}
SRM~\citep{wewer25srm} demonstrates the importance of sequentialization and generation order for reasoning over continuous spatial variables, using predicted uncertainty to select the next position. Patch Forcing~\citep{schusterbauer2026patchforcing} learns per-patch difficulty estimates and adapts denoising progress across an image, allowing easier regions to provide context for harder ones. These approaches motivate adapting generation to the current sample. Our policy is trained against rewards on the completed image. InCoDe~\citep{comas2025incode} instead orders yes or no questions about an existing image to compress it into a short description for a fine-tuned diffusion model. Learned ordering has also been studied in discrete diffusion language models. DCoLT~\citep{huang2025dcolt} uses final-answer rewards to optimize the reverse generation trajectory and its LLaDA generator jointly trains a token-unmasking model and the language model, adapting both generation order and token predictions. \citet{hong2026improving} isolate the ordering component by keeping the language model frozen and learning a position-selection policy from verifiable terminal rewards, with regularization relative to confidence-based reference policies. \citet{fox2026generationorders} approach ordering through variational inference: a posterior conditioned on complete training examples provides a learning signal for an inference-time token selector through an evidence lower bound. 
Concurrent work by \citet{ouyang2026generationorder} learns an unmasking controller with group-relative policy optimization for multimodal diffusion language models. All of these methods operate on discrete tokens and only decide which masked position to reveal next; once revealed, a token stays fixed. Several also rely on quantities that continuous generators with a diffusion head do not provide, such as token confidences or likelihoods. Our work studies continuous-latent diffusion generators and additionally learns which previously generated positions to remove and later regenerate under a rethinking budget.

\paragraph{Revision during generation.}
Allowing accepted predictions to change can correct errors that become apparent as additional context is generated. Path Planning~\citep{peng2025pathplanning} augments masked discrete diffusion sampling with planners that can select both masked and previously unmasked tokens for updating. It also derives an evidence lower bound for training planners with a frozen denoiser. Nemotron-Labs-Diffusion-Image~\citep{li_nemotron-labs-diffusion-image_2026} enables revision of discrete image tokens through an augmented corruption objective and confidence-thresholded token replacement at inference. \method{} learns revision decisions from final-sample rewards without retraining the generator. Its position-selection policy allocates a prescribed removal budget: removed content is excluded from subsequent context and regenerated when its position is selected again. This separates the decision to discard a prediction from its later regeneration.

\paragraph{Reward optimization for generative models.}
DDPO~\citep{black2023ddpo} formulates diffusion denoising as a sequential decision process and optimizes generator parameters using downstream rewards. Flow-GRPO~\citep{liu2025flowgrpo} extends group-relative policy optimization to flow matching models, while Mask-GRPO~\citep{luo2025maskgrpo} develops reward-based post-training for masked text-to-image generators. \method{} applies reinforcement learning to the position-selection policy, leaving content generation to the frozen model. The optimized action probabilities are those of selecting positions to fill or remove, so policy training does not require evaluating the likelihood of generated continuous content.

\paragraph{Distributional objectives.}
FD-loss~\citep{yang2026representationfrechetlossvisual} optimizes representation-space Fr\'echet distance by maintaining a large feature population through a queue or moving moments while backpropagating through the current generated batch. FIDgrad shares the use of evolving population statistics but derives a scalar reward reflecting the influence of a single sample on the FID. It measures the first-order effect of adding probability mass at a generated image's feature. This reward trains the position-selection policy through policy gradients, without differentiating through the generator or feature extractor. It therefore connects a distributional quality objective to discrete generation and revision decisions over continuous content.

\section{Method}

Sequential image generation can be cast as masked image modeling. Starting from an empty canvas, the generator unmasks one or multiple regions (in practice: patches) and fills them by generating suitable pixel values or latents. That process is repeated in autoregressive fashion until the entire canvas has been filled.
The goal of \method{} is to devise a policy in which order to unmask the patches. Additionally, the policy can also choose to re-mask patches, such that they can be regenerated in subsequent steps. The policy is trained to maximize a suitable target metric for the task at hand, e.g., FID~\citep{Heusel2017GANsTB} or Human Preference Score~\citep{wu2023human}.

\subsection{Policy-Guided Masked Generation}
\begin{figure}[h]
    \centering
    \includegraphics[width=1.0\linewidth]{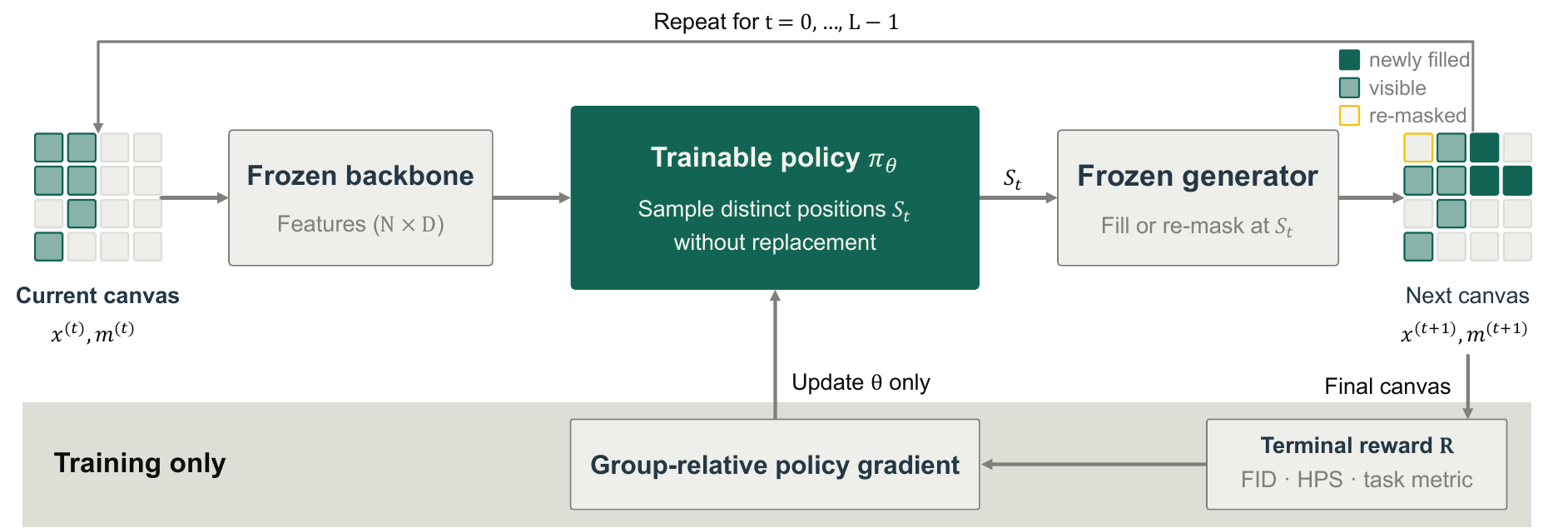}
    \caption{\textbf{Overview of \method{}.}
    The policy processes frozen backbone features using transformer blocks and a linear head, then samples $K_t$ positions without replacement. Selected masked positions are filled, while rethinking masks selected visible positions for later regeneration. A terminal reward trains only the policy through group-relative policy gradients, while the generator remains frozen.}
    \label{fig:ponder_fig1}
\end{figure}

Image generation proceeds for $L$ iterations indexed by $t=0,\ldots,L-1$. Each iteration consists of three steps: evaluating the generator backbone for the current canvas $x^{(t)}$, selecting patches to unmask, and updating the canvas by generating content for the unmasked patches. At each iteration, the input to the backbone is image tokens for the patches that are already visible and MASK tokens for the patches that are still masked. The backbone is evaluated once and computes per-patch features that remain fixed throughout the iteration and constitute the input $h^{(t)}$ for the policy. For diffusion-based approaches, the backbone features are concatenated with the initial noise. Note that the iterations of the sequential generator are different from the ``inner loop'' of denoising steps during diffusion.

The goal of the generation process is to fill all writable patches
$\mathcal{W}\subseteq{\{1,\ldots,N}\}$. A prescribed schedule determines the number $K_t$ of patches to be unmasked in each iteration, while the policy chooses their locations $S_t$. A binary visibility mask $m^{(t)}\in\{0,1\}^{N}$ indicates which positions are already visible as context. Optionally, there may be additional class or text conditioning $c$. Using the context computed at the start of the iteration, the generator fills the selected patches and the mask is updated to the new state:
\begin{equation}
\begin{split}
    x^{(t+1)}_{S_t} &= g_\phi\!\big(x^{(t)},S_t,\epsilon^{(t)}_{S_t},m^{(t)},c\big),\\
    m^{(t+1)}_{S_t}&=\neg m^{(t)}_{S_t}\,
    \label{eq:generator-conditional}
    \end{split}
\end{equation}
where $g_\phi$ denotes the generator and $\epsilon^{(t)}_{S_t}$ is the noise initialization at the selected positions.
Moreover, we extend the policy with a \textit{rethinking} mode. Formally, the selection $S_t$ may contain already visible patches ($m_i^{(t)}=1$), which are then reverted back to the masked state ($m_i^{(t+1)}=0$).

\subsection{Adaptive Patch Selection}

The policy $\pi_\theta$ is a bidirectional transformer that maps the generator features $h^{(t)}$ and start noise vectors to output logits. The policy obtains the same information that the patch-wise diffusion heads obtain. We ablate this choice in Table~\ref{tab:policy-input-ablation}. For MAR-B, we also test the policy with noise that does not match the generator's noise (Appendix~\ref{app:mar-ablations}). The output logits consist of $K_t$ sets of per-patch logits, one for each decision to be taken in the current iteration (Fig.~\ref{fig:ponder_fig1}).  Patches are sampled sequentially, excluding those that have already been selected in the same iteration. 

Let $\mathcal{C}_{t,k}$ denote the candidate patches for decision $k$.
The probability of the ordered selections is
\begin{equation}
    \pi_\theta(a_{t,1:K_t}\mid h^{(t)}) = \prod_{k=1}^{K_t} \pi_{\theta,k}(a_{t,k}\mid h^{(t)},\mathcal{C}_{t,k}).
    \label{eq:iteration-policy}
\end{equation}

All decisions within an iteration share the same generator features. Earlier selections affect later ones only through the candidate mask. After selection, the generator fills $S_t=\{a_{t,1},\ldots,a_{t,K_t}\}$ using the same context from the previous iteration for all of them. Without rethinking, the complete sequence to a complete output visits every writable patch exactly once.

\subsection{\method ing by Revisiting Patches}
\begin{figure}
    \centering
    \includegraphics[width=0.99\linewidth]{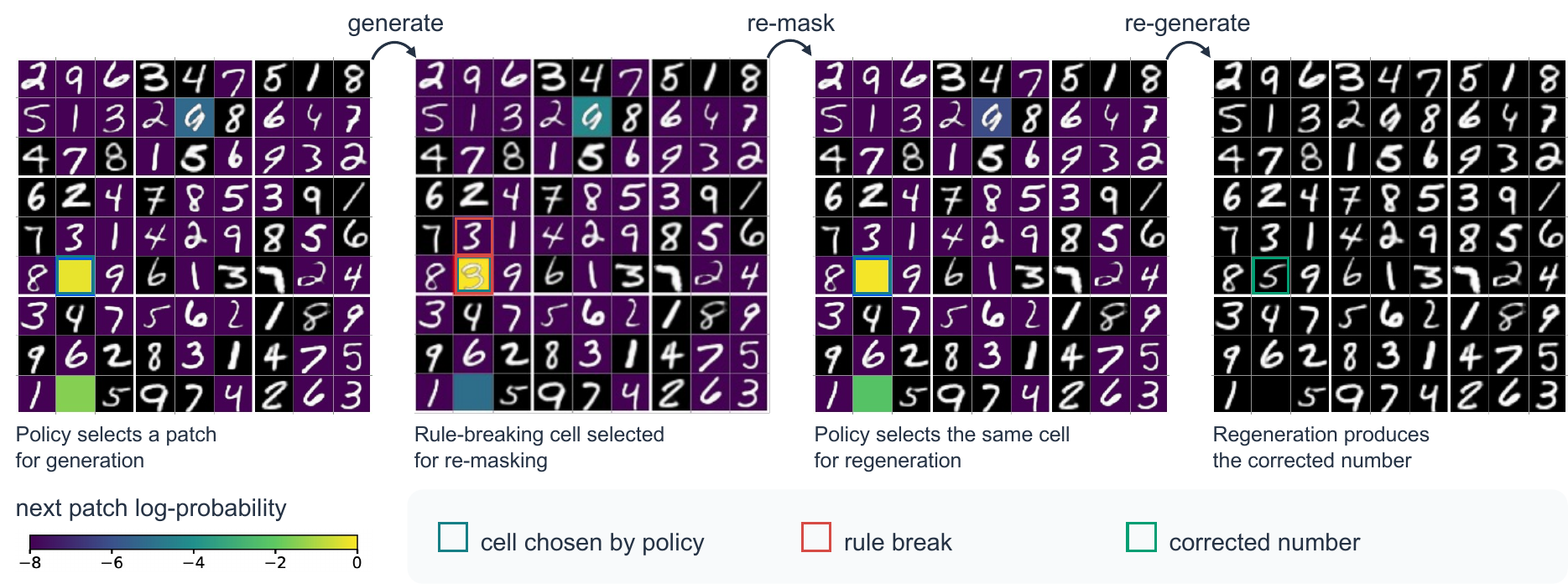}
    \vspace{-0.2cm}
    \caption{An example of sequential generation with rethinking on a visual Sudoku instance. We visualize next patch log probabilities as background color and policy decisions as border colors.
    \vspace{-0.3cm}
    } 
    \label{fig:ponder_thinking}
\end{figure}
Restricting the generation order to strict permutations simplifies the process, but makes it impossible to correct early decisions after additional context becomes available.
To enable such revisions, we allow the policy to remove previously generated content. We note that the same patch selection policy can control both generation and revision. Only a minor change is needed, namely to allow the selection of already visible patches. Whenever a masked patch is selected in iteration $t$, it is filled. When an unmasked patch is selected, it is returned to the masked state, thus excluding it from the backbone evaluation in the next iteration $t+1$. The patch then remains masked until the policy selects it again, at which point it is regenerated using the new context.

Because backbone features are computed before position selection, removal does not change the features used within the current round and requires no additional backbone evaluation in that round. The next scheduled backbone evaluation uses the updated mask.

We prescribe a rethinking budget of $R$ removal actions. Starting from a fully masked canvas, completing the sample requires $N$ net fill actions. Each removal cancels one previous fill and therefore requires one additional fill later in the roll-out. A trajectory with $R$ removals consequently contains $N+R$ fill actions and $R$ removal actions, for a total of
\begin{equation}
T = N + 2R
\label{eq}
\end{equation}
patch decisions. A patch may be removed more than once, with each removal counting toward the budget.
Book-keeping of legal decisions ensures completion: removals are disabled once the budget is exhausted and in the final iteration $L-1$. Conversely, they are forced when necessary to exhaust the budget before the final iteration.

Importantly, rethinking is distinct from repeatedly resampling a position under the same context, because regeneration can happen after other parts of the canvas have been filled, providing more context for the generator. The policy therefore learns not only where to generate next, but also when an earlier prediction is worth revisiting. This differs from fixed refinement schedules that allocate additional computation independently of the evolving sample.

Setting $R=0$ recovers adaptive ordering without rethinking. Because $R$ changes the roll-out horizon but not the frozen generator or the policy parameterization, it can be varied at inference time within the schedules supported by the policy.

\subsection{Group-Relative Policy Optimization}
\label{sec:grpo}
We train the ordering policy from a terminal reward $\mathcal{R}(x^{(L)},c)$ evaluated on the completed sample. For each input and conditioning value, we generate a group of $G$ trajectories. Members of a group share the same initial canvas and generator noise, but sample their policy actions independently, to isolate the effect of the generation order on the final image and the associated reward. 
Let $r_j$ denote the terminal reward of trajectory $j$, and let $\bar r$ and $s$ denote the group mean and population standard deviation. We assign every decision in trajectory $j$ the advantage
\begin{equation}
A_j =
\begin{cases}
(r_j-\bar r)/s, & s>10^{-5},\\
0, & \text{otherwise}.
\end{cases}
\end{equation}
Index the $T=N+2R$ primitive position decisions by $d$. Let $o_{jd}$ contain the policy observation and legal candidate set at decision $d$ of trajectory $j$. At optimizer update $u$, we minimize
\begin{equation}
\mathcal L_u(\theta)
= -\frac{1}{GT}\sum_{j=1}^{G}\sum_{d=1}^{T}
\left[
\operatorname{sg}(A_j)\log\pi_\theta(a_{jd}\mid o_{jd})
\right].
\end{equation}
Here $\operatorname{sg}$ stops gradients. We average this loss across independently collected groups.

Gradients update only the ordering policy, whereas the generator remains frozen.
Batching several independent groups averages the loss across groups.
We use a single policy-gradient update per collected roll-out batch, without reference-policy penalty, importance ratios, clipping, or repeated optimization over the same roll-out.
Further training details are given in Appendix~\ref{app:policy-training}.

\subsection{Distribution-Level Optimization with FIDgrad}
\label{sec:fidgrad}

Standard rewards, like constraint satisfaction or human preference, can be evaluated independently for each generated sample. In contrast, population-level quality metrics, like the Fr\'echet Inception Distance (FID), measure the discrepancy between the distributions of generated and real images and cannot directly be broken down into per-sample rewards. Inspired by the FD-loss~\citep{yang2026representationfrechetlossvisual} for generative model training, we introduce \emph{FIDgrad}: a reward that scores each generated image by its first-order effect on the FID of the current generative distribution. This makes it possible to optimize distribution quality through reinforcement learning, while keeping the generative model frozen.

Let $q$ denote the distribution of Inception features $z=f(x)$ of the generated images, with mean $\mu_g$ and covariance $\Sigma_g$. And let $\mu_r,\Sigma_r$ be the corresponding statistics of the real image data. We work with the standard Gaussian plug-in functional, which depends on $q$ only through its first two moments,
\begin{equation}
\operatorname{FID}(q,p_{\mathrm{real}})=\|\mu_g-\mu_r\|_2^2+\operatorname{tr}(\Sigma_g)
+\operatorname{tr}(\Sigma_r)-2\operatorname{tr}\!\left[\big(\Sigma_r^{1/2}\Sigma_g\Sigma_r^{1/2}\big)^{1/2}\right].
\label{eq:fid}
\end{equation}
To turn Eq.~\ref{eq:fid} into a per-sample quantity, we ask how it responds to placing a small amount of probability mass at a single feature $z$, i.e., we evaluate the
influence function of the functional at $q$.

\begin{proposition}[FIDgrad]
\label{prop:fidgrad}
Assume $\Sigma_g\!\succ\!0$ and $\Sigma_r\!\succ\!0$. For a feature $z$, let
$q_\epsilon\!=\!(1\!-\!\epsilon)q\!+\!\epsilon\delta_z$ with $\delta_z$ a point mass at $z$, and let
$u\!=\!z\!-\!\mu_g$. Then $\epsilon\!\mapsto\!\operatorname{FID}(q_\epsilon,p_{\mathrm{real}})$ is differentiable at $\epsilon=0$ and
\begin{equation}
r_{\mathrm{FIDgrad}}(z)
\,:=\,-\tfrac{\mathrm{d}}{\mathrm{d}\epsilon}
\operatorname{FID}(q_\epsilon,p_{\mathrm{real}})\big|_{\epsilon=0}
\,=\,-2(\mu_g\!-\!\mu_r)^{\!\top}u-u^{\!\top}(I\!-\!B)u+\operatorname{tr}\!\left[(I\!-\!B)\Sigma_g\right],
\label{eq:fidgrad-closed-form}
\end{equation}
where $B=\Sigma_r^{1/2}\!\big(\Sigma_r^{1/2}\Sigma_g\Sigma_r^{1/2}\big)^{-1/2}\Sigma_r^{1/2}$
is symmetric positive definite, and $B=I$ if and only if $\Sigma_g=\Sigma_r$.
\end{proposition}

\begin{wrapfigure}{r}{.5\textwidth}
    \begin{minipage}{\linewidth}
    \centering\captionsetup[subfigure]{justification=centering}
    \includegraphics[width=\linewidth]{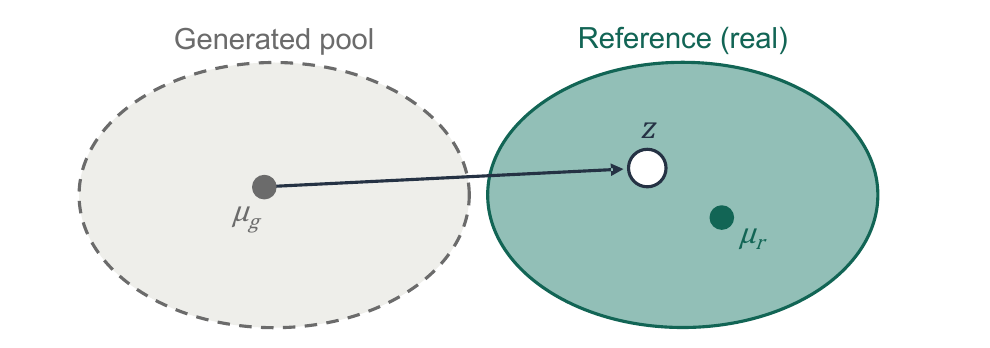}
    \subcaption{Positive mean contribution}
    \label{fig:4a}\par\vfill
    \includegraphics[width=\linewidth]{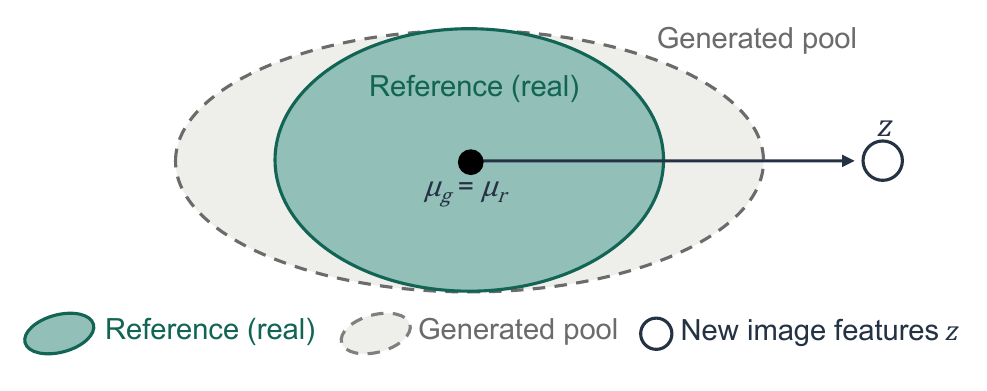}
\vspace{-0.7cm}
    \subcaption{Negative covariance contribution}
    \label{fig:4b}
\end{minipage}
\caption{Illustration of the mean and covariance component of the FIDgrad reward for a new image with features $z$. FID compares Gaussian approximations of generated and reference image distributions in Inception feature space~\citep{Heusel2017GANsTB}. Our reward measures the first order change in FID induced by adding the new image into the pool of generated images. a) The new sample $z$ shifts the mean of the generated distribution towards the reference distribution. This contributes to a positive reward. b) The new sample increases the variance of the generated distribution, although the variance is already higher than the reference distribution. This contributes negatively to the reward. In general a sample affects both: mean and covariance.
\vspace{-1.4cm}
}\label{fig:fidgrad_reward}
\end{wrapfigure}

A positive reward identifies samples whose inclusion would decrease FID to first order. The first term captures the contribution of a sample to correcting the mean of the generated features, the second term captures its contribution to correcting the feature covariance. Figure~\ref{fig:fidgrad_reward} illustrates the resulting effect.

FIDgrad depends on both the current generated distribution and the individual image, so the same image can receive different rewards as the policy evolves. During training, we track the generated mean and covariance using exponential moving first and second moments. Each batch is scored before these estimates are updated, while the real-data moments remain fixed. Initialization and numerical conventions are specified in Appendix \ref{app:fidgrad}. The resulting scores serve as terminal rewards in the group-relative objective.

\section{Experiments}
We train and evaluate \method{} policies for three generative models: MAR-B~\citep[ImageNet,][]{li_autoregressive_2024}, MAR~\citep[Sudoku; similar to SRM,][]{wewer25srm}, and the continuous-token text-to-image model MaskGen-KL-XL~\citep{maskgen}. All generators are frozen throughout our experiments. To speed up policy training, we use DDIM sampling with 25 steps for MaskGen-KL-XL and MAR-B, and flow matching~\citep{lipmanFlowMatchingGenerative2023a} with 12 steps for the Sudoku network. Policy training details are described in Appendix~\ref{app:policy-training}, and generator architectures and inference settings in Appendix~\ref{app:model-parameters}. To our knowledge, \method{} is the first method to learn generation and revision order for continuous-token generators, so no prior method applies directly. We therefore construct baselines that isolate each component.%

\paragraph{Visual Sudoku.}
Following SRM~\citep{wewer25srm}, we generate images of handwritten (MNIST) Sudokus. Such samples can look realistic while violating the Sudoku rules, making the task a test of global constraint satisfaction rather than local image quality. We classify every cell and verify the row, column, and box constraints, using SRM's classifier and its \textit{easy}, \textit{medium}, and \textit{hard} splits, which differ in the number of given cells. The frozen generator is an MAR~\citep{li_autoregressive_2024} with a flow-matching head operating on per-cell VAE latents (Appendix~\ref{app:model-parameters}, Table~\ref{tab:generator-parameters}). A single policy is trained for all difficulty levels with a binary reward for a valid solution (Appendix~\ref{app:sudoku-policy-training}).

Learned ordering alone raises hard-split accuracy from 4.5\% to 56.6\%, well above SRM's uncertainty-based greedy ordering (41.0\%, Table~\ref{tab:sudoku-results}); with a rethinking budget of $R=9$, it reaches 96.7\%. The resampling of patches itself only helps marginally: with random ordering, $R=9$ leaves accuracy at 4.4\%, and with the learned order but randomly chosen removal locations, it reaches only 62.7\% compared to 56.6\% with no rethinking and 96.7\% with learned order and learned rethinking positions. The 34-point gap to learned rethinking is therefore attributable to learning \emph{which} predictions to revisit. The policy trained with $R=9$ also generalizes to larger test-time budgets, reaching 98.6\% at $R=128$ (Figure~\ref{fig:sudoku-test-time-scaling}).

\begin{table}[t]
    \centering
    
    \vspace{0.5em}
    
    \begin{minipage}[t]{0.62\linewidth}
    \vspace{0pt}
    \centering
    \scriptsize
    \setlength{\tabcolsep}{2.5pt}
    \renewcommand{\arraystretch}{1.08}
    \begin{tabular*}{\linewidth}{@{\extracolsep{\fill}}lrrrrr@{}}
        \toprule
        \multicolumn{6}{@{}c@{}}{\textbf{MAR-Sudoku}} \\
        \midrule
        Position policy & $R$ & Overall $\uparrow$ & Hard $\uparrow$ & Medium $\uparrow$ & Easy $\uparrow$ \\
        \midrule
        Random order & 0 & 52.7 & 4.5 & 56.9 & 96.7 \\
        SRM greedy   & 0 & 71.6 & 41.0 & 77.2 & 96.8 \\
        Random order & 9 & 52.7 & 4.4 & 57.0 & 96.7 \\
        \midrule
        \method{}      & 0 & 80.2 & 56.6 & 87.1 & 96.9 \\
        \method{} w. random rethinking & 9 & 83.1 & 62.7 & 89.0 & 97.6 \\
        \method{} & 9 & \textbf{98.2} & \textbf{96.7} & \textbf{98.4} & \textbf{99.6} \\
        \bottomrule
    \end{tabular*}
    \captionsetup{font=footnotesize,skip=2pt,justification=raggedright,singlelinecheck=false}
    \captionof{table}{
        Exact-solution accuracy (\%) on MNIST Sudoku. \(R\) is the rethinking budget. The two \(R=9\) \method{} rows retain learned order and the same removal budget, varying only the selection of removal locations.
    }
    \label{tab:sudoku-results}
    \end{minipage}\hfill
    \begin{minipage}[t]{0.35\linewidth}
        \vspace{0pt}
        \centering
        \pgfplotstableread[col sep=comma]{plots/sudoku_scaling.csv}\sudokuscalingdata
\pgfplotsset{
    discard sudoku difficulty/.style={
        x filter/.code={
            \pgfplotstablegetelem{\coordindex}{difficulty}\of\sudokuscalingdata
            \edef\sudokudifficulty{\pgfplotsretval}%
            \def\targetdifficulty{#1}%
            \ifx\sudokudifficulty\targetdifficulty\else
                \def\pgfmathresult{nan}%
            \fi
        }
    }
}
\begin{tikzpicture}
    \begin{axis}[
        scaling axis,
        ylabel={Hard accuracy (\%) $\uparrow$},
        xmin=-0.15, xmax=7.15,
        xtick={0,2.3219280949,4.0874628413,6.0223678130,7.0112272554},
        xticklabels={0,4,16,64,128},
        ymin=52, ymax=101,
        ytick={60,80,100},
    ]
        \addplot[ponder scaling,discard sudoku difficulty=hard, color=gray] table[
            x expr={ln(\thisrow{rethinking_steps}+1)/ln(2)},
            y expr={100*\thisrow{accuracy}},
        ]{\sudokuscalingdata};
    \end{axis}
\end{tikzpicture}
        \captionsetup{font=footnotesize,skip=2pt,justification=raggedright,singlelinecheck=false}
        
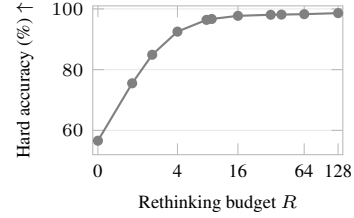
\captionof{figure}{Scaling the rethinking budget at test time improves accuracy dramatically.}
        \label{fig:sudoku-test-time-scaling}
    \end{minipage}
    \vspace{-1.em}

\end{table}

\paragraph{ImageNet (MAR).}
We apply \method{} to class-conditional ImageNet generation with the frozen MAR-B generator, training the policy with FIDgrad (Sec.~\ref{sec:fidgrad}). Since FIDgrad is computed in Inception feature space, we additionally report FDr-5 to check against reward hacking. It follows the FDr-6 metric of \citet{yang2026representationfrechetlossvisual} but excludes the Inception backbone, and thus measures Fr\'echet distance in five representation spaces that are not used by the reward (Appendix~\ref{app:mar-scaling}). Without rethinking, \method{} reduces FID from 4.09 to 2.78 and increases IS from 181.22 to 206.55 compared with random ordering. FDr-5 improves as well (13.69 $\to$ 13.28). Increasing the test-time rethinking budget of the same policy to $R=32$ further improves all three metrics, reaching 2.70 FID, 12.47 FDr-5, and 212.60 IS (Figure~\ref{fig:mar-test-time-scaling}). Additional budgets are reported in Appendix Table~\ref{tab:mar-imagenet-scaling}. Over three seeds of continued training, learned rethinking improves FDr-5 over random rethinking on every seed, while the difference in FID is within seed variation (Appendix~\ref{app:mar-ablations}).
Figure~\ref{fig:mar-test-time-scaling} plots the corresponding FDr-5 trend.
\begin{table}[t]
    \centering

    \begin{minipage}[t]{0.62\linewidth}
    \vspace{0pt}
    \centering
    \scriptsize
    \setlength{\tabcolsep}{3pt}
    \renewcommand{\arraystretch}{1.08}
    \begin{tabular*}{\linewidth}{@{\extracolsep{\fill}}lrrrr@{}}
        \toprule
        \multicolumn{5}{@{}c@{}}{\textbf{ImageNet: MAR-B}} \\
        \midrule
        Position policy & $R$ & FID $\downarrow$ & FDr-5 $\downarrow$ & IS $\uparrow$ \\
        \midrule
        Random & 0 & 4.09 & 13.69 & 181.22 \\
        Random & 32 & 3.80 & 12.82 & 185.90 \\
        \midrule
        \method & 0 & 2.78 & 13.28 & 206.55 \\
        \method{} w. random rethinking & 32 & 2.76 & 12.68 & 209.00 \\
        \method & 32 & \textbf{2.70} & \textbf{12.47} & \textbf{212.60} \\

        \bottomrule
    \end{tabular*}
    \captionsetup{font=footnotesize,skip=2pt,justification=raggedright,singlelinecheck=false}
    \captionof{table}{ImageNet-256 results with frozen MAR-B.}
    \label{tab:mar-imagenet-results}

    \end{minipage}\hfill
    \begin{minipage}[t]{0.35\linewidth}
        \vspace{0pt}
        \centering
        \begin{tikzpicture}
    \begin{axis}[
        scaling axis,
        ylabel={FDr-5 $\downarrow$},
        xmin=-0.1, xmax=5.15,
        xtick={0,1.5849625007,3.1699250014,5.0443941194},
        xticklabels={0,2,8,32},
        ymin=12.35, ymax=13.4,
        ytick={12.4,12.8,13.2},
    ]
        \addplot[ponder scaling, color=gray] table[
            col sep=comma, x=plot_x, y=fdr5,
        ]{plots/mar_scaling.csv};
    \end{axis}
\end{tikzpicture}
        \captionsetup{font=footnotesize,skip=2pt,justification=raggedright,singlelinecheck=false}
        
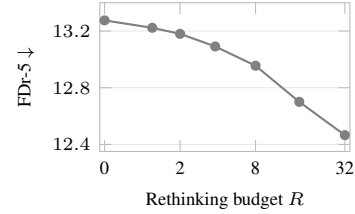
\captionof{figure}{MAR-B test-time scaling; FDr-5 is not used as reward.}
        \label{fig:mar-test-time-scaling}
    \end{minipage}

    \vspace{-0.5em}

\end{table}

\begin{figure}
    \centering
\includegraphics[width=1\linewidth]{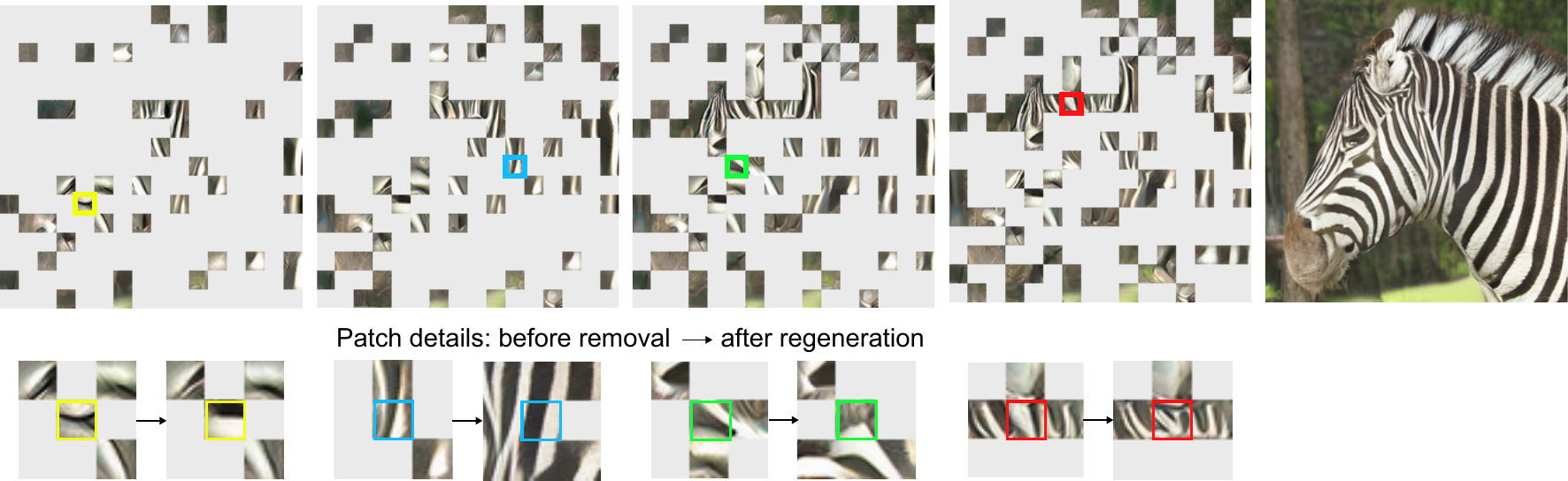}

    \caption{We visualize the successive generation of a MAR-B image. In the bottom row we visualize patches before they have been remasked and after the same patch has been re-generated.}
    \label{fig:mar_zebra_rollout}
\end{figure}
\paragraph{Text-to-image generation.
}
We train two separate ordering policies for the frozen MaskGen-KL-XL generator. One uses Human-Preference-Score v2.1~\citep{wuHumanPreferenceScore2023} (HPSv2) as reward on HPSv2 training prompts. We train it with a uniformly sampled rethinking budget from $R=0$ to $R=8$; each update uses one sampled budget for all trajectories. We evaluate on the official HPSv2 benchmark prompts and additionally report PickScore~\citep{kirstainPickaPicOpenDataset2023} and ImageReward~\citep{xuImageRewardLearningEvaluating2023}.
At $R=0$, \method{} scores 26.50 HPSv2 versus 25.05 for random ordering. Increasing $R$ to 4 with the same policy raises HPSv2 to 26.95 and ImageReward from 0.68 to 0.73. At $R=32$, HPSv2 reaches 27.80 and ImageReward reaches 0.81 (Figure~\ref{fig:maskgen-preference-scaling}).
We train a separate policy with the same $R\in\{0,\ldots,8\}$ mixture on the GenEval-style training prompts released by Flow-GRPO~\citep{liu2025flowgrpo}, using its dense partial-credit reward. We report the original binary GenEval metric~\citep{ghoshGenEvalObjectFocusedFramework2023}. Results are shown in Tables~\ref{tab:maskgen-hps-results} and~\ref{tab:maskgen-geneval-results}.
On GenEval, overall accuracy rises from 37.3\% with random ordering to 41.1\% with learned ordering at $R=0$. At $R=4$, the learned policy scores 42.6\% versus 38.0\% for random ordering (Table~\ref{tab:maskgen-geneval-results}). Figure~\ref{fig:maskgen-geneval-scaling} shows the learned-policy test-time scaling trend.

\begin{table}[t]
    \centering
    \begin{minipage}[t]{0.62\linewidth}
    \vspace{0pt}
    \centering
    \scriptsize
    \setlength{\tabcolsep}{3pt}
    \renewcommand{\arraystretch}{1.08}
    \begin{tabular*}{\linewidth}{@{\extracolsep{\fill}}lrrrr@{}}
        \toprule
        \multicolumn{5}{@{}c@{}}{\textbf{MaskGen-KL-XL}} \\
        \midrule
        Position policy & $R$ & HPSv2 $\uparrow$ & PickScore $\uparrow$ & ImageReward $\uparrow$ \\
        \midrule
        Random & 0 & 25.05 & 21.21 & 0.55 \\
        Random & 4 & 25.19 & 21.26 & 0.59 \\
        \midrule
        \method & 0 & 26.50 & 21.32 & 0.68 \\
        \method{} w. random rethinking & 4 & 26.40 & 21.33 & 0.67 \\
        \method & 4 & \textbf{26.95} & \textbf{21.41} & \textbf{0.73} \\
        
        \bottomrule
    \end{tabular*}
    \captionsetup{font=footnotesize,skip=2pt,justification=raggedright,singlelinecheck=false}
    \captionof{table}{
        MaskGen-KL-XL preference scores on the official HPSv2 prompts.
    \label{tab:maskgen-hps-results}
    }
    \end{minipage}\hfill
    \begin{minipage}[t]{0.35\linewidth}
        \vspace{0pt}
        \centering
        \begin{tikzpicture}
    \begin{axis}[
        scaling axis,
        ylabel={ImageReward $\uparrow$},
        xmin=-0.1, xmax=5.15,
        xtick={0,1.5849625007,3.1699250014,5.0443941194},
        xticklabels={0,2,8,32},
        ymin=0.65, ymax=0.85,
        ytick={0.65,0.70,0.75,0.80,0.85},
        legend style={
            draw=none, fill=none, font=\tiny,
            at={(0.03,0.97)}, anchor=north west,
        },
    ]
        \addplot[ponder scaling, color=gray] table[
            col sep=comma, x=plot_x, y=ponder_imagereward
        ]{plots/maskgen_preference_scaling.csv};
    \end{axis}
\end{tikzpicture}
        \captionsetup{font=footnotesize,skip=2pt,justification=raggedright,singlelinecheck=false}
        
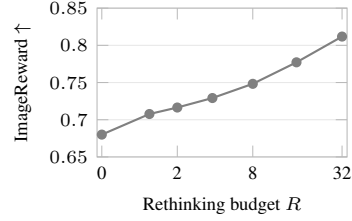
\captionof{figure}{MaskGen test time scaling for the HPSv2 policy. }
        \label{fig:maskgen-preference-scaling}
    \end{minipage}
\end{table}

\begin{table}[t]
    \centering

    \begin{minipage}[t]{0.62\linewidth}
    \vspace{0pt}
    \centering
    \scriptsize
    \setlength{\tabcolsep}{1.7pt}
    \renewcommand{\arraystretch}{1.08}
    \begin{tabular*}{\linewidth}{@{\extracolsep{\fill}}lrrrrrrrr@{}}
        \toprule
        \multicolumn{9}{@{}c@{}}{\textbf{MaskGen-KL-XL}} \\
        \midrule
        Policy & $R$ & Overall $\uparrow$ & Single $\uparrow$ & Two $\uparrow$ & Count $\uparrow$ & Color $\uparrow$ & Pos. $\uparrow$ & Attr. $\uparrow$ \\
        \midrule
        Random  & 0 & 37.3 & 88.4 & 32.1 & 21.3 & 66.5 & 5.0 & 10.8 \\
        Random  & 4 & 38.0 & 89.7 & 37.6 & 21.6 & 63.8 & 5.3 & 9.8 \\
        \midrule
        \method & 0 & 41.1 & 93.4 & 35.1 & 30.9 & \textbf{67.8} & 6.8 & 12.5 \\
        \makecell[l]{\method{}\\ w. random rethinking}
                & 4 & 41.9 & \textbf{95.0} & \textbf{39.1} & 29.7 & 65.2 & \textbf{7.5} & 15.0 \\
        \method & 4 & \textbf{42.6} & \textbf{95.0} & 38.6 & \textbf{31.9} & 67.0 & \textbf{7.5} & \textbf{15.5} \\
        \bottomrule
    \end{tabular*}
    \captionsetup{font=footnotesize,skip=2pt,justification=raggedright,singlelinecheck=false}
    \captionof{table}{
        MaskGen-KL-XL results on the official GenEval prompts. Scores are percentages.
    }
    \label{tab:maskgen-geneval-results}
    \end{minipage}\hfill
    \begin{minipage}[t]{0.35\linewidth}
        \vspace{0pt}
        \centering
        \begin{tikzpicture}
    \begin{axis}[
        scaling axis,
        ylabel={GenEval overall (\%) $\uparrow$},
        xmin=-0.1, xmax=5.15,
        xtick={0,1.5849625007,3.1699250014,5.0443941194},
        xticklabels={0,2,8,32},
        ymin=40, ymax=45,
        ytick={40,42,44},
        legend style={
            draw=none, fill=none, font=\tiny,
            at={(0.97,0.03)}, anchor=south east,
        },
        legend cell align=left,
    ]
        \addplot[ponder scaling, color=gray] table[
            col sep=comma, x=plot_x, y=overall,
        ]{plots/maskgen_geneval_scaling_latest.csv};
    \end{axis}
\end{tikzpicture}
        \captionsetup{font=footnotesize,skip=2pt,justification=raggedright,singlelinecheck=false}
        
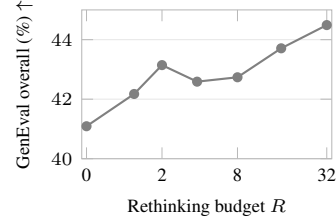
\captionof{figure}{MaskGen test time scaling for the GenEval policy.}
        \label{fig:maskgen-geneval-scaling}
    \end{minipage}
    \vspace{-.5em}

\end{table}

\vspace{-0.2em}
\begin{table}[th]
    \centering
    \small
    \setlength{\tabcolsep}{4pt}
    \renewcommand{\arraystretch}{1.08}
    \caption{Policy-input ablation with frozen MaskGen-KL-L and MAR-Sudoku. The benefit of observing generation noise depends on the generator and task: it helps MaskGen in this comparison, but is not required for strong Sudoku  performance. Generator features remain useful in both settings.}
    \label{tab:policy-input-ablation}
    \begin{tabular}{@{}lrrrrrr@{}}
        \toprule
        & \multicolumn{2}{c}{MaskGen-KL-L}
        & \multicolumn{4}{c}{MAR-Sudoku ($R=9$)} \\
        & \multicolumn{2}{c}{HPSv2 $\uparrow$}
        & \multicolumn{4}{c}{Accuracy (\%) $\uparrow$} \\
        \cmidrule(lr){2-3}\cmidrule(l){4-7}
        Policy input & $R=0$ & $R=4$ & Overall & Hard & Medium & Easy \\
        \midrule
        Generator features + noise & \textbf{26.11} & \textbf{26.61}
            & 98.21
            & 96.68
            & 98.39
            & \textbf{99.57} \\
        Generator features only & 25.01 & 25.71
            & \textbf{98.59}
            & \textbf{97.55}
            & \textbf{98.64}
            & \textbf{99.57} \\
        Latent state only & 25.05 & 25.42
            & 63.01
            & 26.92
            & 63.59
            & 98.53 \\
        \bottomrule
    \end{tabular}

\end{table}

\vspace{-0.5em}

\section{Conclusion and Discussion}

We present \method{}, a general framework for learning the generation order of frozen masked generative models. Rather than relying on fixed, random, or locally greedy schedules, \method{} learns a position-selection policy from rewards defined on the completed sample. The same policy can also revisit earlier predictions through \emph{rethinking}, allowing decisions made with limited context to be removed and reconsidered once additional information becomes available. Across visual constraint solving, class-conditional image generation, and text-to-image synthesis, this improves the observed metrics over the generator's default sampling schedule, while leaving the underlying generator untouched. Importantly, improvements are not limited to the reward we used. This suggests that the observed gains are not solely due to overoptimization of the training reward. Generation order thus becomes an optimizable component of the generative process rather than a predefined heuristic.

\paragraph{Limitations.} The policy adds a forward pass per generation iteration. Without rethinking, this increases sampling time by 2--12\%, and rethinking can require additional generator rounds (Appendix~\ref{app:runtime}). Its behavior is strongly dependent on the reward that is used for training. Reward design therefore remains a critical requirement for policy success. Finally, rethinking is governed by a prescribed budget and not allocated on demand. Learning \emph{when} to stop, in addition to where to generate and revise, and extending the action space to inference parameters such as guidance strength, are natural next steps toward generative models that learn their own inference strategy.

\paragraph{Acknowledgments.} This work was funded, in part, by the Max Planck ETH Center for Learning Systems and Huawei Technologies Oy (Finland) Co. Ltd.

\subsection*{AI use statement}

We used generative AI tools to assist with implementing and debugging research code, provide feedback on experimental methodology, identify relevant literature, assist with the mathematical derivations for the FIDgrad reward and revise the manuscript. The authors reviewed and revised the AI-assisted text, checked references against the original sources, checked the mathematical derivations step by step under the stated assumptions and verified the resulting reward expression and reviewed and tested AI-assisted code. We take responsibility for the final manuscript, methods, results, and accompanying code, including all content produced with generative AI assistance.

\bibliography{iclr2026_conference}
\bibliographystyle{iclr2026_conference}

\newpage

\appendix
\section{Appendix}
\setcounter{totalnumber}{6}
\setcounter{topnumber}{4}
\setcounter{bottomnumber}{3}
\renewcommand{\topfraction}{0.9}
\renewcommand{\bottomfraction}{0.8}
\renewcommand{\textfraction}{0.07}
\renewcommand{\floatpagefraction}{0.8}
\setlength{\floatsep}{8pt plus 2pt minus 2pt}
\setlength{\intextsep}{8pt plus 2pt minus 2pt}
\setlength{\textfloatsep}{12pt plus 2pt minus 4pt}

\subsection{Policy training details}
\label{app:policy-training}

\subsubsection{Shared setup}

The generators remain frozen while we train their ordering policies.

Trajectories within a conditioning group share generation noise, indexed by position and the number of previous fills. The policy observes the noise for each position's next fill, while policy actions are sampled using a separate random stream. Evaluations use matched noise across rethinking budgets, with the noise capacity held fixed.

All ordering policies are bidirectional pre-norm Transformer encoders with
GELU activations, learned positional embeddings, and zero dropout. Unless
otherwise noted, each position's input concatenates the frozen generator
feature with the Gaussian noise for the next fill. Table~\ref{tab:policy-parameters}
reports the trainable policy parameters and optimization settings.

\begin{table}[th]
    \centering
    \small
    \renewcommand{\arraystretch}{1.08}
    \caption{Dynamic-removal policy architectures and optimization settings.}
    \label{tab:policy-parameters}
    \begin{tabularx}{\textwidth}{@{}l *{3}{>{\centering\arraybackslash}X}@{}}
        \toprule
        Parameter & MAR-B & MaskGen-KL-XL & MAR-Sudoku \\
        \midrule
        \multicolumn{4}{@{}l}{\textit{Architecture}} \\
        Number of positions & 256 & 32 & 81 \\
        Generator feature width & 768 & 1280 & 256 \\
        Noise width & 16 & 16 & 32 \\
        Input projection & $784\!\rightarrow\!768$ & $1296\!\rightarrow\!1024$ & $288\!\rightarrow\!256$ \\
        Transformer depth & 6 & 4 & 2 \\
        Model width & 768 & 1024 & 256 \\
        Attention heads & 8 & 8 & 4 \\
        Feed-forward width & 3072 & 4096 & 1024 \\
        Action heads & 7 & 1 & 2 \\
        Trainable parameters & 43.334M & 51.749M & 1.675M \\
        \midrule
        \multicolumn{4}{@{}l}{\textit{Optimization}} \\
        Optimizer & Adam & Adam & Adam \\
        Learning rate & $2\!\times\!10^{-5}$ & $2\!\times\!10^{-5}$ & $2\!\times\!10^{-5}$ \\
        Learning-rate schedule & constant, then linear decay & constant & constant \\
        Adam $(\beta_1,\beta_2)$ & $(0.9,0.95)$ & $(0.9,0.95)$ & $(0.9,0.95)$ \\
        Maximum training updates & 50k & 50k & 50k \\
        Reported checkpoint (updates)
            & \leavevmode11k
            & \leavevmode11k (HPSv2), 4.1k (GenEval)
            & \leavevmode47k \\
        Conditioning inputs/update & 128 & 128 & 128 \\
        Trajectories per group & 8 & 8 & 16 \\
        Trajectories per update & 1024 & 1024 & 2048 \\
        Gradient-norm clip & 1.0 & 1.0 & 1.0 \\
        \bottomrule
    \end{tabularx}
\end{table}

All policies use Adam with $\epsilon=10^{-8}$ and zero weight decay.
During policy training, both the policy and frozen generator use bfloat16
mixed precision. Each update contains 128 independent conditioning inputs,
each expanded into one group of trajectories. Evaluation precision is
reported with each experiment.

\subsubsection{MAR-B}
\label{app:fidgrad}

We train the MAR-B policy with the FIDgrad reward and a rethinking budget of
$R=8$. The learning rate is held constant for 2,000 updates and then decays
linearly to $2\times10^{-6}$ over 2,000 updates.

At update $u$, let $(\mu_u,\Sigma_u)$ denote the generated feature moments
and $(\mu_r,\Sigma_r)$ the fixed real moments. For $d=z-\mu_u$, the
perturbation $q_\eta=(1-\eta)q+\eta\delta_z$ changes the mean and covariance
to first order by $d$ and $dd^\top-\Sigma_u$, respectively. Combining these
with the derivatives of FID gives the reward in
Eq.~\ref{eq:fidgrad-closed-form} under the assumptions of
Proposition~\ref{prop:fidgrad}.

\paragraph{Estimated moments.}
We maintain the generated mean $\mu_u$ and raw second moment $M_u$,
with $\Sigma_u=\operatorname{sym}(M_u-\mu_u\mu_u^\top)$.
Every trajectory is scored using the current moments before they are
updated. One trajectory is selected uniformly from each group of eight
to estimate the batch moments. For selected features $z_1,\ldots,z_n$,
the updates are
\begin{align}
 \hat\mu_u &= n^{-1}\sum_{i=1}^n z_i,&
 \hat M_u &= n^{-1}\sum_{i=1}^n z_i z_i^\top,\\
 \mu_{u+1} &= (1-\alpha)\mu_u+\alpha\hat\mu_u,&
 M_{u+1} &= (1-\alpha)M_u+\alpha\hat M_u,
\end{align}
with $\alpha=0.001$. The moments are initialized from 50,048 separately
generated images. Rewards use 2048-dimensional Inception features from
the FD-Loss implementation~\citep{yang2026representationfrechetlossvisual}
and fixed ImageNet-256 real reference statistics. Moment and matrix
computations use double precision. Inverse square roots are computed by
eigendecomposition, using a pseudoinverse on numerically singular directions.

\paragraph{Interpretation of the update.}
The moving moment estimates and within-group reward normalization yield
a surrogate objective for policy training. The resulting updates need
not decrease population FID at every step.

\paragraph{Training cost.}
\label{app:training-cost}
Each MAR-B training trajectory is a complete 64-round sampling run with
25-step DDIM sampling of all generated and regenerated tokens and VAE
decoding, which costs 6.3~TFLOP, about $14\times$ the per-image cost of a
MAR-B training step. The reported policy (update 11,000) was trained on
11.3M trajectories, without training images and without gradients through
the generator. This amounts to $1.4\times10^{20}$~FLOPs, about 30\% of the
$4.5\times10^{20}$~FLOPs of MAR-B's 800-epoch pretraining (with cached VAE
latents, excluding tokenizer training). About two thirds of this compute
is spent on collecting trajectories (generator unroll and policy forward
passes) and one third on the policy update; the reward is negligible.
Training took about 115 hours including periodic 50k-sample FID
evaluations, on four RTX PRO 6000 GPUs. We did not optimize training efficiency: each
batch of rollouts is used for a single on-policy update and every round is
replayed for the gradient. Reusing rollouts across several clipped updates,
replaying a subset of rounds or using a smaller policy would reduce this
cost without changing the generator. FLOPs count matrix multiplications,
convolutions and attention, measured with PyTorch's FLOP counter.

\subsubsection{MaskGen-KL-XL}
\label{app:maskgen-training}

We train separate policies against HPSv2.1 and the dense GenEval reward, using the same architecture and optimization settings in Table~\ref{tab:policy-parameters}.
The 553 official GenEval prompts (four images per prompt) are disjoint from the Flow-GRPO training prompts. We report the
original binary GenEval score on these prompts.

\subsubsection{MAR-Sudoku}
\label{app:sudoku-policy-training}

We train the MAR-Sudoku policy with a rethinking budget of $R=9$ and a terminal reward of 1 for a correct completed Sudoku and 0 otherwise.
Training starts from partially filled grids. The number of given cells follows a triangular distribution with maximum density at zero and zero density at 81; their locations are sampled randomly.

\subsection{Generator architectures and inference settings}
\label{app:model-parameters}

Table~\ref{tab:generator-parameters} summarizes the generator architectures and inference settings.
Generator parameter counts include the contextual backbone and shared generative head, but exclude tokenizers/VAEs, the MaskGen text encoder, and the Sudoku reward classifier. Both MaskGen-KL-XL policies use the architecture shown here. 

\begin{table}[tbp]
    \centering
    \small
    \renewcommand{\arraystretch}{1.08}
    \caption{Generator architectures and dynamic-removal inference settings.}
    \label{tab:generator-parameters}
    \begin{tabularx}{\textwidth}{@{}l *{3}{>{\centering\arraybackslash}X}@{}}
        \toprule
        Parameter & MAR-B & MaskGen-KL-XL & MAR-Sudoku \\
        \midrule
        \multicolumn{4}{@{}l}{\textit{Architecture}} \\
        Tokenizer & KL16 VAE & TA-TiTok-BL32 VAE & MNIST VAE \\
        Token layout & $16\!\times\!16$ grid & 32-token sequence & $9\!\times\!9$ grid \\
        Number of tokens & 256 & 32 & 81 \\
        Token dimension & 16 & 16 & 32 \\
        Backbone & encoder--decoder & MM-DiT decoder & encoder--decoder \\
        Model width & 768 & 1280 & 256 \\
        Encoder depth & 12 & --- & 3 \\
        Decoder/joint depth & 12 & 20 & 3 \\
        Attention heads & 12 & 16 & 4 \\
        Feed-forward width & 3072 & 5120 & 1024 \\
        Generative head & DiffLoss & DiffLoss & FlowMatch \\
        Head depth & 6 & 8 & 3 \\
        Head width & 1024 & 1280 & 256 \\
        Frozen parameters & 207.925M & 1237.345M & 6.215M \\
        \midrule
        \multicolumn{4}{@{}l}{\textit{Inference}} \\
        Sampler & DDIM ($\eta=0$) & DDIM ($\eta=0$) & Euler \\
        Denoising steps & 25 & 25 & 12 \\
        CFG schedule & constant $1.0$ & revealed-token linear & none \\
        Parallelization & 64-round cosine & sequential & 2 actions/round, then 1 \\
        Selections per round & 1--7 & 1 & 1--2 \\
        Rounds & 64 & 40 & 81 (empty grid) \\
        Removal budget & 8 & 4 & 9 \\
        Position decisions & 272 & 40 & 99 (empty grid) \\
        \bottomrule
    \end{tabularx}
\end{table}

For MAR-B, let $C_r$ be the cumulative number of position decisions after
round $r$.  The cosine schedule is
\[
C_0=0,\qquad
C_r=\max\!\left\{C_{r-1}+1,
\left\lceil272\left[1-\cos\!\left(\frac{\pi r}{128}\right)\right]\right\rceil
\right\},
\]
with the final cumulative count fixed at $C_{64}=272$.

MaskGen uses classifier-free
guidance scale $s=1+2[32-\max(1,M)]/32$, where $M$ is the number of masked
tokens.

For MAR-Sudoku with $g$ clues, let $W=81-g$ denote the number of writable
cells. A rethinking budget of $R$ requires $D=W+2R$ position
decisions: $W+R$ fills and $R$ removals. If round $t$ makes $K_t$
decisions and the trajectory has $L$ rounds, then
\[
    \sum_{t=1}^{L} K_t = D.
\]
All $K_t$ decisions in one round share the same generator features, giving
$L$ backbone evaluations. For Sudoku,
$K_t\in\{1,2\}$. When $W\geq 2R+1$, using $K_t=2$ for $2R$ rounds and
$K_t=1$ for the remaining $W-2R$ rounds gives $L=W$. Thus, on an empty
grid with the training budget $R=9$, the policy makes 99 position decisions
over 81 backbone evaluations.

MAR-B and MaskGen-KL-XL use released pretrained weights without fine-tuning.
MAR-Sudoku is trained for 500,000 updates with AdamW,
$(\beta_1,\beta_2)=(0.9,0.95)$, $\epsilon=10^{-8}$, weight decay $0.2$, and a
cosine learning-rate schedule from $8\!\times\!10^{-4}$ to
$1\!\times\!10^{-4}$.

\subsection{Sudoku test-time scaling}
\label{app:sudoku-scaling-runtime}

Table~\ref{tab:sudoku-scaling-results} reports accuracy
across rethinking budgets.

\begin{table}[t]
    \centering
    \small
    \setlength{\tabcolsep}{5pt}
    \renewcommand{\arraystretch}{1.08}
    \caption{
        Test-time scaling of the \method{} policy trained with $R=9$ on MNIST
        Sudoku. Overall and hard-split exact-solution accuracy are measured on
        30,000 held-out boards balanced across difficulty levels.
    }
    \label{tab:sudoku-scaling-results}
    \begin{tabular}{@{}rrrrr@{}}
        \toprule
        $R$ & Overall (\%) $\uparrow$ & Hard (\%) $\uparrow$ & Medium (\%) $\uparrow$ & Easy (\%) $\uparrow$ \\
        \midrule
         0 & 80.17 & 56.60 & 87.07 & 96.85 \\
         1 & 89.68 & 75.50 & 94.22 & 99.33 \\
         2 & 93.62 & 84.90 & 96.50 & 99.46 \\
         4 & 96.53 & 92.52 & 97.53 & 99.53 \\
         8 & 98.07 & 96.39 & 98.30 & 99.52 \\
         9 & 98.21 & 96.68 & 98.39 & \textbf{99.57} \\
        16 & 98.56 & 97.72 & 98.60 & 99.37 \\
        32 & 98.68 & 98.07 & 98.60 & 99.36 \\
        40 & 98.72 & 98.13 & 98.57 & 99.47 \\
        64 & 98.75 & 98.27 & \textbf{98.70} & 99.27 \\
       128 & \textbf{98.89} & \textbf{98.61} & 98.64 & 99.41 \\
        \bottomrule
    \end{tabular}
    
\end{table}

At $R=9$, accuracy is 98.21\%, 98.02\%, and 97.94\% across three evaluation
seeds of the same policy, compared with 80.17\%, 79.96\%, and 80.29\% at
$R=0$. For the first seed, the improvement is 18.04 percentage points.

\subsection{MAR-B test-time scaling}
\label{app:mar-scaling}
Table~\ref{tab:mar-imagenet-scaling} reports additional \method{} budgets
together with random-policy baselines.
\begin{table}[th]
    \centering
    \small
    \setlength{\tabcolsep}{5.5pt}
    \renewcommand{\arraystretch}{1.08}
    \caption{
        Class-conditional ImageNet results with the frozen MAR-B generator.
        The FIDgrad policy is trained with a rethinking budget of $R=8$ and
        evaluated at every listed test-time budget $R$. We generate 50,000
        images (50 per class) with DDIM using 25 denoising steps and guidance
        scale 1.
    }
    \label{tab:mar-imagenet-scaling}
    \begin{tabular}{@{}lrrrrr@{}}
        \toprule
        Position policy & $R$ & Decisions & FID $\downarrow$ & FDr-5 $\downarrow$ & IS $\uparrow$ \\
        \midrule
        Random & 0 & 256 & 4.09 & 13.69 & 181.22 \\
        Random & 8 & 272 & 3.97 & 13.42 & 182.22 \\
        Random & 32 & 320 & 3.80 & 12.82 & 185.90 \\
        \midrule
        \method & 0 & 256 & 2.78 & 13.28 & 206.55 \\
        \method & 1 & 258 & 2.78 & 13.22 & 206.13 \\
        \method & 2 & 260 & 2.78 & 13.18 & 207.06 \\
        \method & 4 & 264 & 2.75 & 13.09 & 207.78 \\
        \method & 8 & 272 & 2.74 & 12.96 & 209.48 \\
        \method & 16 & 288 & 2.72 & 12.70 & 212.05 \\
        \method & 32 & 320 & 2.70 & 12.47 & 212.60 \\
        \bottomrule
    \end{tabular}
\end{table}

\paragraph{FDr-5 evaluation.}
FDr-5 is the mean of the normalized Fr\'echet distances in ConvNeXt, DINOv2,
MAE, SigLIP, and CLIP representations, excluding the Inception reward space,
using the FD-Loss evaluator~\citep{yang2026representationfrechetlossvisual}.

\subsection{MaskGen preference test-time scaling}
\label{app:maskgen-preference-scaling}
At test time, the policy uses one position decision per round, giving
$32+2R$ rounds for $R\in\{0,1,2,4,8,16,32\}$.

Training uses the official HPSv2 training prompts for the reward; evaluation
uses all 3,200 HPSv2 benchmark prompts.
All learned-policy budgets use the same start noise.  We report HPSv2, PickScore, and ImageReward (Table~\ref{tab:maskgen-hps-scaling}).
\begin{table}[th]
    \centering
    \small
    \setlength{\tabcolsep}{5pt}
    \renewcommand{\arraystretch}{1.08}
    \caption{
        Preference test-time scaling with frozen MaskGen-KL-XL and the
        update-11,000 HPSv2 policy. All scores use the 3,200 official HPSv2
        prompts, DDIM-25, float32, and seed 42. Each budget uses $32+2R$
        position decisions, one per round.
    }
    \label{tab:maskgen-hps-scaling}
    \begin{tabular}{@{}rrrrr@{}}
        \toprule
        $R$ & Decisions & HPSv2 $\uparrow$ & PickScore $\uparrow$ & ImageReward $\uparrow$ \\
        \midrule
         0 & 32 & 26.50 & 21.32 & 0.68 \\
         1 & 34 & 26.66 & 21.36 & 0.71 \\
         2 & 36 & 26.79 & 21.38 & 0.72 \\
         4 & 40 & 26.95 & 21.41 & 0.73 \\
         8 & 48 & 27.17 & 21.44 & 0.75 \\
        16 & 64 & 27.47 & 21.49 & 0.78 \\
        32 & 96 & \textbf{27.80} & \textbf{21.54} & \textbf{0.81} \\
        \bottomrule
    \end{tabular}
\end{table}

\subsection{Inference runtime}
\label{app:runtime}

Tables~\ref{tab:runtime-mar}--\ref{tab:runtime-sudoku} report sampling time
per image on one RTX~3090~Ti with our evaluation code. Without rethinking,
the policy increases sampling time by 12\% (MAR-B), 2\% (MaskGen-KL-XL) and
9\% (MAR-Sudoku). The cost of rethinking depends on how many decisions fit
into one generator round. MAR-B and MAR-Sudoku fill several positions per
round, and their action heads (Table~\ref{tab:policy-parameters}) let
removals share rounds with fills, so rethinking adds no generator rounds up
to $R=8$ for MAR-B and $R=40$ for MAR-Sudoku. MaskGen-KL-XL takes one
decision per round, following the generator's sequential sampler, so each
removal and its refill add one round each ($32+2R$ rounds) and time grows
linearly with $R$. Predicting several decisions per round, as for MAR-B,
would reduce this cost.
For MAR-B, Table~\ref{tab:runtime-mar} also gives random order the policy's
additional time as additional generator rounds: at the runtime of \method{}
at $R=0$ and $R=32$, random order with 72 and 85 rounds reaches 4.08 and
4.02 FID.

\begin{table}[th]
    \centering
    \caption{
        MAR-B inference cost on one RTX~3090~Ti at batch size 32. Rounds are
        generator backbone evaluations per sample. Numbers in the plot give rounds.
    }
    \label{tab:runtime-mar}
    \begin{minipage}[c]{0.55\linewidth}
    \centering
    \scriptsize
    \setlength{\tabcolsep}{3pt}
    \renewcommand{\arraystretch}{1.08}
    \begin{tabular*}{\linewidth}{@{\extracolsep{\fill}}lrrrrrr@{}}
        \toprule
        Position policy & $R$ & Rounds & \makecell[r]{Time $\downarrow$\\(ms)}
            & FID $\downarrow$ & FDr-5 $\downarrow$ & IS $\uparrow$ \\
        \midrule
        Random & 0 & 64 & 263 & 4.09 & 13.69 & 181.22 \\
        Random & 0 & 72 & 295 & 4.08 & 13.63 & 179.64 \\
        Random & 0 & 85 & 347 & 4.02 & 13.50 & 181.19 \\
        Random & 32 & 76 & 309 & 3.80 & 12.82 & 185.90 \\
        \midrule
        \method{} & 0 & 64 & 293 & 2.78 & 13.28 & 206.55 \\
        \method{} & 8 & 64 & 292 & 2.74 & 12.96 & 209.48 \\
        \method{} & 16 & 68 & 310 & 2.72 & 12.70 & 212.05 \\
        \method{} & 32 & 76 & 345 & \textbf{2.70} & \textbf{12.47} & \textbf{212.60} \\
        \bottomrule
    \end{tabular*}
    \end{minipage}\hfill
    \begin{minipage}[c]{0.42\linewidth}
    \centering
    \begin{tikzpicture}
        \begin{axis}[
            runtime axis,
            xmin=250, xmax=365,
            xtick={250,300,350},
            ylabel={FID $\downarrow$},
            ymin=2.5, ymax=4.45,
            ytick={2.8,3.2,3.6,4.0},
            yticklabel style={/pgf/number format/fixed, /pgf/number format/fixed zerofill,
                /pgf/number format/precision=1},
            legend style={at={(0.98,0.45)}, anchor=east},
        ]
            \addplot[random scaling, nodes near coords, point meta=explicit symbolic,
                visualization depends on={value \thisrow{anchor} \as \labelanchor},
                every node near coord/.append style={anchor=\labelanchor}]
                table[meta=label] {
                    x y label anchor
                    262.6 4.09 64 {south east}
                    295.3 4.08 72 south
                    346.8 4.02 85 {south west}
                };
            \addlegendentry{Random}
            \addplot[baselineorange, only marks, mark=square*, mark size=1.5pt,
                nodes near coords, point meta=explicit symbolic, forget plot,
                every node near coord/.append style={anchor=west, xshift=1pt}]
                coordinates {(309.4,3.80)[$R$=32]};
            \addplot[ponder scaling]
                coordinates {(293.2,2.78) (292.5,2.74) (310.2,2.72) (344.6,2.70)};
            \addlegendentry{\method{}}
        \end{axis}
    \end{tikzpicture}
    \end{minipage}
\end{table}

\begin{table}[th]
    \centering
    \caption{MaskGen-KL-XL inference cost on one RTX~3090~Ti at batch size 16.}
    \label{tab:runtime-maskgen}
    \begin{minipage}[c]{0.55\linewidth}
    \centering
    \scriptsize
    \setlength{\tabcolsep}{3pt}
    \renewcommand{\arraystretch}{1.08}
    \begin{tabular*}{\linewidth}{@{\extracolsep{\fill}}lrrrr@{}}
        \toprule
        Position policy & $R$ & Rounds & \makecell[r]{Time $\downarrow$\\(ms)}
            & HPSv2 $\uparrow$ \\
        \midrule
        Random & 0 & 32 & 472 & 25.05 \\
        Random & 4 & 40 & 588 & 25.19 \\
        Random & 32 & 96 & 1395 & 25.58 \\
        \midrule
        \method{} & 0 & 32 & 479 & 26.50 \\
        \method{} & 4 & 40 & 595 & 26.95 \\
        \method{} & 8 & 48 & 712 & 27.17 \\
        \method{} & 16 & 64 & 945 & 27.47 \\
        \method{} & 32 & 96 & 1411 & \textbf{27.80} \\
        \bottomrule
    \end{tabular*}
    \end{minipage}\hfill
    \begin{minipage}[c]{0.42\linewidth}
    \centering
    \begin{tikzpicture}
        \begin{axis}[
            runtime axis,
            xmin=400, xmax=1500,
            xtick={500,1000,1500},
            ylabel={HPSv2 $\uparrow$},
            ymin=24.2, ymax=28.1,
            ytick={24,25,26,27,28},
            legend style={at={(0.98,0.52)}, anchor=east},
        ]
            \addplot[random scaling]
                coordinates {(471.7,25.05) (588.2,25.19) (1394.8,25.58)};
            \addlegendentry{Random}
            \addplot[ponder scaling]
                coordinates {(479.0,26.50) (595.3,26.95) (711.6,27.17) (945.4,27.47) (1411.2,27.80)};
            \addlegendentry{\method{}}
        \end{axis}
    \end{tikzpicture}
    \end{minipage}
\end{table}

\begin{table}[th]
    \centering
    \caption{MAR-Sudoku inference cost on one RTX~3090~Ti at batch size 2048.}
    \label{tab:runtime-sudoku}
    \begin{minipage}[c]{0.55\linewidth}
    \centering
    \scriptsize
    \setlength{\tabcolsep}{3pt}
    \renewcommand{\arraystretch}{1.08}
    \begin{tabular*}{\linewidth}{@{\extracolsep{\fill}}lrrrr@{}}
        \toprule
        Position policy & $R$ & Rounds & \makecell[r]{Time $\downarrow$\\(ms)}
            & Accuracy (\%) $\uparrow$ \\
        \midrule
        Random & 0 & 81 & 5.68 & 52.7 \\
        Random & 9 & 81 & 5.75 & 52.7 \\
        Random & 40 & 81 & 5.81 & 54.7 \\
        Random & 64 & 105 & 7.57 & 56.0 \\
        Random & 128 & 169 & 12.29 & 57.8 \\
        \midrule
        \method{} & 0 & 81 & 6.20 & 80.2 \\
        \method{} & 9 & 81 & 6.40 & 98.2 \\
        \method{} & 40 & 81 & 6.50 & 98.7 \\
        \method{} & 64 & 105 & 8.50 & 98.8 \\
        \method{} & 128 & 169 & 13.88 & \textbf{98.9} \\
        \bottomrule
    \end{tabular*}
    \end{minipage}\hfill
    \begin{minipage}[c]{0.42\linewidth}
    \centering
    \begin{tikzpicture}
        \begin{axis}[
            runtime axis,
            xmin=5, xmax=14.5,
            xtick={6,8,10,12,14},
            ylabel={Accuracy (\%) $\uparrow$},
            ymin=45, ymax=103,
            ytick={50,60,70,80,90,100},
            legend style={at={(0.98,0.45)}, anchor=east},
        ]
            \addplot[random scaling]
                coordinates {(5.679,52.70) (5.752,52.68) (5.811,54.74) (7.572,55.95) (12.293,57.79)};
            \addlegendentry{Random}
            \addplot[ponder scaling]
                coordinates {(6.203,80.17) (6.396,98.21) (6.501,98.72) (8.501,98.75) (13.877,98.89)};
            \addlegendentry{\method{}}
        \end{axis}
    \end{tikzpicture}
    \end{minipage}
\end{table}

\subsection{MAR-B ablations}
\label{app:mar-ablations}

\paragraph{Seed variability.}
Table~\ref{tab:mar-seed-variability} continues training the reported policy
with three seeds. Learned rethinking improves FDr-5 over random rethinking
at the same budget on every seed ($-0.19\pm0.02$, $p=0.005$), whereas the
FID difference ($-0.02\pm0.05$) is within the variation across seeds.

\paragraph{Policy noise.}
Table~\ref{tab:mar-policy-noise} evaluates the reported policy with
independently drawn noise as its input, while the generator keeps its own
noise. The policy then loses its advantage over random order, so the trained
policy relies on the noise input to choose positions. Each position's noise
is drawn once before sampling and is not changed by the policy. The policy
therefore still only selects the order, but is better informed about what
each fill will produce. Since the
policy never observed mismatched noise during training, this evaluation
does not show how much of the gain a policy trained without noise retains.

\paragraph{Fine-tuning on \method{} orders.}
Since \method{}'s orders depend on the sample, there is no single learned
order to train a generator on. Instead, we record the orders in which the
reported policy generates ImageNet training images when the ground-truth
patches are filled in instead of the generator's samples (teacher forcing),
and fine-tune MAR-B on the resulting masks for 12.5k steps (20 epochs) with
MAR's training recipe. Evaluated with the policy at $R=0$
(Table~\ref{tab:mar-finetune}), training on \method{} orders improves FDr-5
from 13.28 to 12.70, IS from 206.55 to 207.88, and FID marginally from 2.78
to 2.75. Training or fine-tuning policy and generator jointly is left for
future work.

\begin{table}[th]
    \centering
    \small
    \setlength{\tabcolsep}{5pt}
    \renewcommand{\arraystretch}{1.08}
    \caption{
        MAR-B variability over three seeds (mean $\pm$ standard deviation).
        \method{} rows: the reported policy (update 11,000) is trained for
        1,000 further updates with seeds 43, 44, and 45 and evaluated at
        update 12,000 with the same seed; the policies are not trained from
        scratch. Random rows: evaluation seeds 43--45. A paired two-sided
        $t$-test between \method{} and \method{} w.\ random rethinking at
        $R=32$ gives $p=0.005$ for FDr-5, $p=0.60$ for FID, and $p=0.25$ for IS.
    }
    \label{tab:mar-seed-variability}
    \begin{tabular}{@{}lrrrr@{}}
        \toprule
        Position policy & $R$ & FID $\downarrow$ & FDr-5 $\downarrow$ & IS $\uparrow$ \\
        \midrule
        Random & 0 & 4.10 $\pm$ 0.04 & 13.71 $\pm$ 0.05 & 179.12 $\pm$ 0.99 \\
        Random & 32 & 3.83 $\pm$ 0.06 & 12.82 $\pm$ 0.01 & 186.27 $\pm$ 0.34 \\
        \midrule
        \method{} & 0 & 2.88 $\pm$ 0.02 & 13.37 $\pm$ 0.22 & 206.38 $\pm$ 1.38 \\
        \method{} w.\ random rethinking & 32 & 2.83 $\pm$ 0.05 & 12.80 $\pm$ 0.20 & 211.66 $\pm$ 1.08 \\
        \method{} & 32 & 2.81 $\pm$ 0.03 & 12.61 $\pm$ 0.21 & 213.10 $\pm$ 1.64 \\
        \bottomrule
    \end{tabular}
\end{table}

\begin{table}[th]
    \centering
    \small
    \setlength{\tabcolsep}{5pt}
    \renewcommand{\arraystretch}{1.08}
    \caption{
        Policy-noise ablation with MAR-B at $R=0$. The trained \method{}
        policy observes either the start noise that the generator uses or
        independently drawn noise; the generator always uses its own noise.
        Evaluation seed 42.
    }
    \label{tab:mar-policy-noise}
    \begin{tabular}{@{}llrrr@{}}
        \toprule
        Position policy & Policy noise input & FID $\downarrow$ & FDr-5 $\downarrow$ & IS $\uparrow$ \\
        \midrule
        \method{} & Generator noise & 2.78 & 13.28 & 206.55 \\
        \method{} & Independent noise & 4.09 & 13.72 & 179.25 \\
        \midrule
        Random & -- & 4.09 & 13.69 & 181.22 \\
        \bottomrule
    \end{tabular}
\end{table}

\begin{table}[th]
    \centering
    \small
    \setlength{\tabcolsep}{4pt}
    \renewcommand{\arraystretch}{1.08}
    \caption{
        Fine-tuning MAR-B on \method{} generation orders. MAR-B training is
        resumed for 20 epochs with MAR's training recipe, masking either
        uniformly at random (control) or with the masks of \method{}
        trajectories. Both variants are evaluated at $R=0$ with random order
        and with the unchanged \method{} policy. Evaluation seed 42.
    }
    \label{tab:mar-finetune}
    \begin{tabular}{@{}lrrrrrr@{}}
        \toprule
        & \multicolumn{3}{c}{Random order} & \multicolumn{3}{c}{\method{}} \\
        \cmidrule(lr){2-4}\cmidrule(l){5-7}
        Generator & FID $\downarrow$ & FDr-5 $\downarrow$ & IS $\uparrow$
            & FID $\downarrow$ & FDr-5 $\downarrow$ & IS $\uparrow$ \\
        \midrule
        MAR-B & 4.09 & 13.69 & 181.22 & 2.78 & 13.28 & 206.55 \\
        Fine-tuned, random masks & 4.12 & 13.82 & 178.49 & 2.77 & 13.36 & 205.15 \\
        Fine-tuned, \method{} masks & 3.86 & 12.99 & 183.88 & 2.75 & 12.70 & 207.88 \\
        \bottomrule
    \end{tabular}
\end{table}

\subsection{Qualitative examples}
\label{app:qualitative}

\begin{figure}[!htbp]
    \centering
    \begin{tabular}{@{}cc@{}}
        \small Random ordering ($R=0$) & \small \method{} ($R=4$) \\[-0.2em]
        \includegraphics[width=0.42\textwidth]{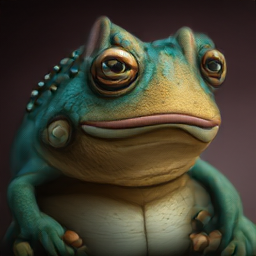} &
        \includegraphics[width=0.42\textwidth]{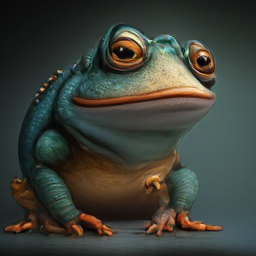} \\[-0.35em]
        \multicolumn{2}{c}{\parbox{0.9\textwidth}{\centering\footnotesize\emph{A hyper-realistic representation of the hypnotoad from Futurama.}}} \\[0.3em]
        \includegraphics[width=0.42\textwidth]{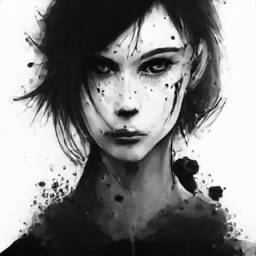} &
        \includegraphics[width=0.42\textwidth]{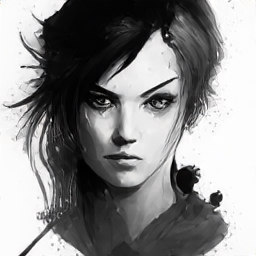} \\[-0.35em]
        \multicolumn{2}{c}{\parbox{0.9\textwidth}{\centering\footnotesize\emph{A goth anime woman with a symmetrical and attractive face in a black and white watercolor headshot art on ArtStation.}}} \\[0.3em]
        \includegraphics[width=0.42\textwidth]{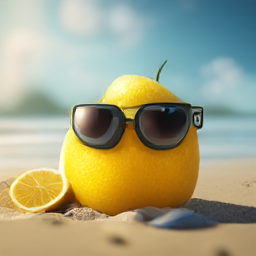} &
        \includegraphics[width=0.42\textwidth]{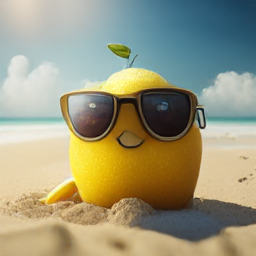} \\[-0.35em]
        \multicolumn{2}{c}{\parbox{0.9\textwidth}{\centering\footnotesize\emph{A Pixar lemon wearing sunglasses on a beach.}}}
    \end{tabular}
    \captionsetup{font=footnotesize,skip=2pt}
    \caption{\textbf{MaskGen-KL-XL examples.} Random ordering at $R=0$ (left) versus the HPSv2 policy at $R=4$ (right), using the same prompts and DDIM-25 sampler.}
    \label{fig:maskgen-qualitative}

\end{figure}

\begin{figure}[!htbp]
    \centering
    \begin{tabular}{@{}cc@{}}
        \small Random ordering ($R=0$) & \small \method{} ($R=4$) \\[-0.2em]
        \includegraphics[width=0.42\textwidth]{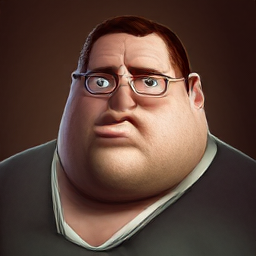} &
        \includegraphics[width=0.42\textwidth]{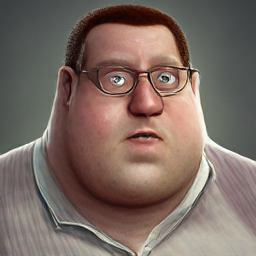} \\[-0.35em]
        \multicolumn{2}{c}{\parbox{0.9\textwidth}{\centering\footnotesize\emph{The image depicts Peter Griffin, a cartoon character from the television show ``Family Guy.''}}} \\[0.3em]
        \includegraphics[width=0.42\textwidth]{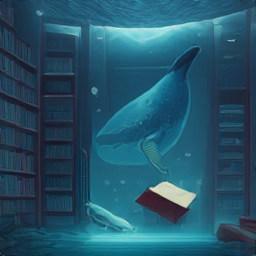} &
        \includegraphics[width=0.42\textwidth]{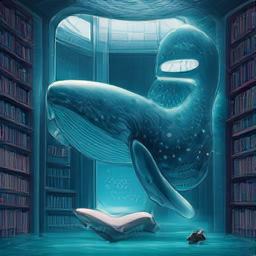} \\[-0.35em]
        \multicolumn{2}{c}{\parbox{0.9\textwidth}{\centering\footnotesize\emph{A whale is reading a book about avoiding Japanese spears in an underwater library.}}}
    \end{tabular}
    \captionsetup{font=footnotesize,skip=2pt}
    \caption{\textbf{Additional MaskGen-KL-XL examples.} Random ordering at $R=0$ (left) versus the HPSv2 policy at $R=4$ (right), using the same prompts and DDIM-25 sampler.}
    \label{fig:maskgen-qualitative-extra}

\end{figure}

\begin{figure}[!htbp]
    \centering
    \begin{tabular}{@{}c@{\hspace{0.01\textwidth}}c@{\hspace{0.03\textwidth}}c@{\hspace{0.01\textwidth}}c@{}}
        \multicolumn{2}{c}{\small Panda} & \multicolumn{2}{c}{\small Goldfish} \\
        \includegraphics[width=0.225\textwidth]{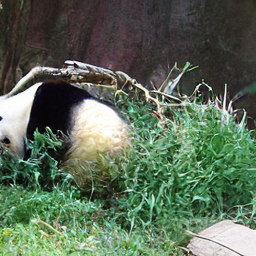} &
        \includegraphics[width=0.225\textwidth]{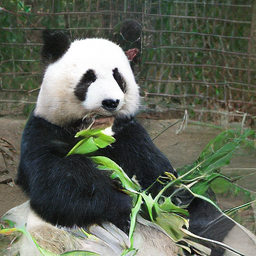} &
        \includegraphics[width=0.225\textwidth]{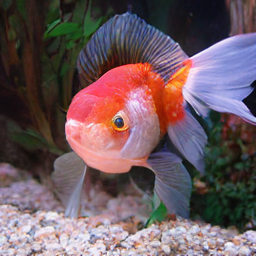} &
        \includegraphics[width=0.225\textwidth]{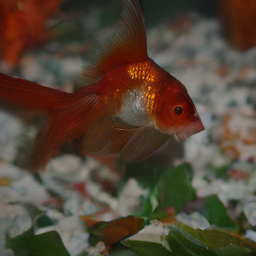} \\
        \scriptsize Random & \scriptsize \method{} & \scriptsize Random & \scriptsize \method{} \\[0.6em]
        \multicolumn{2}{c}{\small Husky} & \multicolumn{2}{c}{\small Yellow bird} \\
        \includegraphics[width=0.225\textwidth]{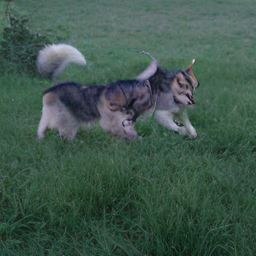} &
        \includegraphics[width=0.225\textwidth]{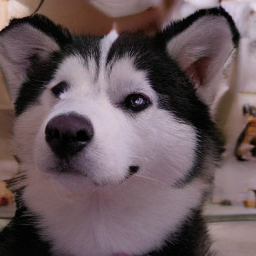} &
        \includegraphics[width=0.225\textwidth]{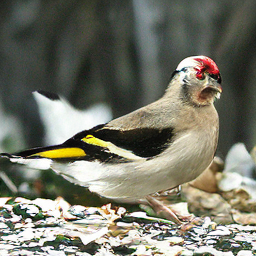} &
        \includegraphics[width=0.225\textwidth]{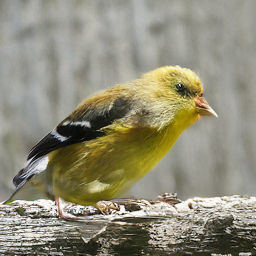} \\
        \scriptsize Random & \scriptsize \method{} & \scriptsize Random & \scriptsize \method{}
    \end{tabular}
    \caption{\textbf{MAR-B ImageNet qualitative examples.} Random-policy and \method{} samples with a rethinking budget of $R=8$. Each class is shown as a random/\method{} pair.}
    \label{fig:mar-qualitative}
\end{figure}

\begin{figure}[!htbp]
    \centering
    \includegraphics[width=\linewidth]{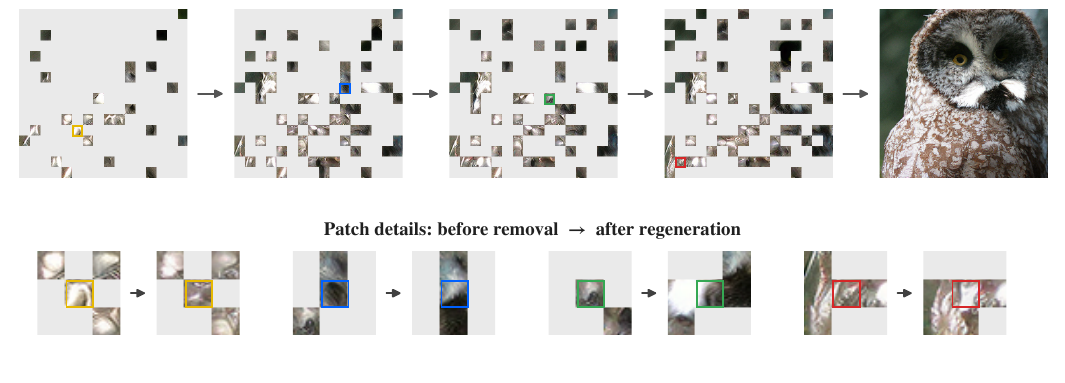}\\[0.5em]
    \vspace{-1em}
    
    \caption{\textbf{MAR-B rethinking trajectory.} Each upper sequence shows successive partial canvases and the final image; the lower strip enlarges patches before removal and after regeneration. Colored boxes identify the revised locations.}
    \label{fig:mar-rethinking-trajectories}
\end{figure}

Figures~\ref{fig:maskgen-qualitative} and~\ref{fig:maskgen-qualitative-extra} compare random-order and policy-guided MaskGen-KL-XL samples. Figure~\ref{fig:mar-qualitative} shows MAR-B ImageNet samples, and Figure~\ref{fig:mar-rethinking-trajectories} illustrates patch removal and later regeneration during MAR-B sampling.


\end{document}